\pdfoutput=1 
\documentclass[journal,two side,web]{ieeecolor}
\usepackage{generic}
\usepackage{cite}
\usepackage{amsmath,amssymb,amsfonts}
\usepackage{algorithmic}
\usepackage{graphicx}
\usepackage{hyperref}
\hypersetup{hidelinks=true}
\usepackage{textcomp}
\def\BibTeX{{\rm B\kern-.05em{\sc i\kern-.025em b}\kern-.08em
    T\kern-.1667em\lower.7ex\hbox{E}\kern-.125emX}}
\usepackage{sidecap}
\usepackage{framed}

\usepackage{amssymb}
\usepackage{latexsym}

\usepackage{url}
\usepackage{xcolor}

\definecolor{newcolor}{rgb}{.8,.349,.1}

\usepackage{cite}
\usepackage{amsmath,amssymb,amsfonts}
\usepackage{graphicx}
\usepackage{textcomp}
\def\BibTeX{{\rm B\kern-.05em{\sc i\kern-.025em b}\kern-.08em
    T\kern-.1667em\lower.7ex\hbox{E}\kern-.125emX}}

\usepackage{times}  
\usepackage{helvet}  
\usepackage{courier}  
\usepackage{url}  
\usepackage{graphicx}  
\usepackage{float}
\usepackage{multirow}
\usepackage{diagbox}
\usepackage{amsmath}
\usepackage[linesnumbered,titlenumbered,ruled,vlined,commentsnumbered,noend]{algorithm2e}

\usepackage{times}
\usepackage{soul}
\usepackage[utf8]{inputenc}
\usepackage[font=small,labelfont=bf]{caption}
\usepackage{makecell}
\usepackage{comment}
\usepackage{color}

\usepackage{epsfig}
\usepackage{graphicx}
\usepackage{amsmath}
\usepackage{amssymb}

\usepackage{url}
\usepackage{multirow}
\usepackage{moresize}
\usepackage{epstopdf}
\usepackage{array}

\usepackage[switch]{lineno}  %
\usepackage{booktabs}
\usepackage{pifont}
\usepackage{wrapfig}
\usepackage{subfigure}
\usepackage{dsfont}

\def\argmax{\operatornamewithlimits{arg\,max}}

\def\BEFPGD{$\mathrm{BEF_{PGD}}$}
\def\CBEFPGD{$\mathrm{CBEF_{PGD}}$}

\usepackage{multirow}
\usepackage{colortbl}

\usepackage{xspace}

\usepackage{pifont}
\newcommand{\revised}[1]{{#1}}
\newcommand{\revisedd}[1]{{#1}}

\definecolor{royalblue}{RGB}{65,105,225}

\usepackage{colortbl}
\definecolor{tabgray}{rgb}{0.85,0.85,0.85}
\definecolor{top1}{rgb}{1.0, 0.6, 0.6} 
\definecolor{top2}{rgb}{0.98, 0.91, 0.71}
\definecolor{top3}{rgb}{0.91, 1.0, 1.0}
\definecolor{top1-2}{rgb}{1.0, 0.66, 0.66} 
\definecolor{top1-3}{rgb}{1.0, 0.72, 0.72} 
\definecolor{top1-4}{rgb}{1.0, 0.78, 0.78} 
\definecolor{top1-5}{rgb}{1.0, 0.84, 0.84} 
\definecolor{top1-6}{rgb}{1.0, 0.90, 0.90} 
\definecolor{top1-7}{rgb}{1.0, 0.96, 0.96}

\makeatletter
\DeclareRobustCommand\onedot{\futurelet\@let@token\@onedot}
\def\@onedot{\ifx\@let@token.\else.\null\fi\xspace}
\def\eg{\emph{e.g}\onedot}

\def\etal{\emph{et al}\onedot}
\makeatother

\definecolor{darkorange}{rgb}{1.0, 0.55, 0.0}
\definecolor{lincolngreen}{rgb}{0.11, 0.35, 0.02}
\definecolor{cornflowerblue}{rgb}{0.39, 0.58, 0.93}
\definecolor{cobalt}{rgb}{0.0, 0.28, 0.67}
\definecolor{coralred}{RGB}{255,127,0}

\usepackage{scalerel}
\usepackage{tikz}
\usetikzlibrary{svg.path}
\definecolor{orcidlogocol}{HTML}{A6CE39}
\tikzset{
  orcidlogo/.pic={
    \fill[orcidlogocol] svg{M256,128c0,70.7-57.3,128-128,128C57.3,256,0,198.7,0,128C0,57.3,57.3,0,128,0C198.7,0,256,57.3,256,128z};
    \fill[white] svg{M86.3,186.2H70.9V79.1h15.4v48.4V186.2z}
                 svg{M108.9,79.1h41.6c39.6,0,57,28.3,57,53.6c0,27.5-21.5,53.6-56.8,53.6h-41.8V79.1z M124.3,172.4h24.5c34.9,0,42.9-26.5,42.9-39.7c0-21.5-13.7-39.7-43.7-39.7h-23.7V172.4z}
                 svg{M88.7,56.8c0,5.5-4.5,10.1-10.1,10.1c-5.6,0-10.1-4.6-10.1-10.1c0-5.6,4.5-10.1,10.1-10.1C84.2,46.7,88.7,51.3,88.7,56.8z};
  }
}
\newcommand\orcidicon[1]{\href{https://orcid.org/#1}{\mbox{\scalerel*{
\begin{tikzpicture}[yscale=-1,transform shape]
\pic{orcidlogo};
\end{tikzpicture}
}{|}}}}

\begin{document}



\title{Adversarial Exposure Attack on \\Diabetic Retinopathy Imagery Grading}%

\author{
{Yupeng~Cheng$^*$\,\orcidicon{0000-0003-3865-2947}\,,~\IEEEmembership{Member,~IEEE}, Qing~Guo$^{*\dag}$\,\orcidicon{0000-0003-0974-9299}\,,~\IEEEmembership{Member,~IEEE}, Felix~Juefei-Xu\,\orcidicon{0000-0002-0857-8611}\,,~\IEEEmembership{Member,~IEEE}, 
Huazhu~Fu,\orcidicon{0000-0002-9702-5524}\,,~\IEEEmembership{Senior~Member,~IEEE}, 
Shang-Wei~Lin\,\orcidicon{0000-0002-9726-3434}\,,~\IEEEmembership{Member,~IEEE},
Weisi~Lin\,\orcidicon{0000-0001-9866-1947}\,,~\IEEEmembership{Fellow,~IEEE}
}
\thanks{
Manuscript received September 1, 2023; revised July 25, 2024 and September
21, 2024; accepted September 21, 2024. Date of current version September 21, 2024. This research is supported by the National Research Foundation, Singapore, and DSO National Laboratories under the AI Singapore Programme (AISG Award No: AISG2-GC-2023-008), Career Development Fund (CDF) of Agency for Science, Technology and Research (A*STAR) (No.: C233312028), and National Research Foundation, Singapore and Infocomm Media Development Authority under its Trust Tech Funding Initiative (No. DTC-RGC-04). (Corressponding author: Qing Guo.)\\
Yupeng Cheng (E-mail: ycheng024@e.ntu.edu.sg), Weisi Lin (E-mail: wslin@ntu.edu.sg) are with Nanyang Technological University, Singapore. \\
Qing Guo$^\dag$ (corresponding author. E-mail: tsingqguo@ieee.org) is with Institute of High Performance Computing (IHPC) and Centre for Frontier AI Research (CFAR), Agency for Science, Technology and Research (A*STAR), Singapore.\\
Felix Juefei-Xu (E-mail: juefei.xu@gmail.com) is with New York University, USA. \\
Huazhu Fu (E-mail: hzfu@ieee.org) is with Institute of High Performance Computing, Agency for Science, Technology and Research, Singapore.\\
Shang-Wei Lin (E-mail: shangwei.lin@singaporetech.edu.sg) is with Singapore Institute of Technology, Singapore. 
}
}

\maketitle

\begin{abstract}
Diabetic Retinopathy (DR) is a leading cause of vision loss around the world. To help diagnose it, numerous cutting-edge works have built powerful deep neural networks (DNNs) to automatically grade DR via retinal fundus images (RFIs).
However, RFIs are commonly affected by camera exposure issues that may lead to incorrect grades.
The mis-graded results can potentially pose high risks to an aggravation of the condition.
\revised{
In this paper, we study this problem from the viewpoint of adversarial attacks.
We identify and introduce a novel solution to an entirely new task, termed as \textit{adversarial exposure attack}, which is able to produce natural exposure images and mislead the state-of-the-art DNNs.
}
We validate our proposed method on a real-world public DR dataset with three DNNs, \eg, ResNet50, MobileNet, and EfficientNet, demonstrating that our method achieves high image quality and success rate in transferring the attacks.
Our method reveals the potential threats to DNN-based automatic DR grading and would benefit the development of exposure-robust DR grading methods in the future. 
\end{abstract}




\section{Introduction}\label{sec:intro}



Diabetic Retinopathy (DR) is the leading cause of vision impairment and blindness among working-age adults globally \cite{lee2015epidemiology,ShuWeiTing2017}. DR is an eye disease associated with diabetes and, if detected and graded in time, its progression to vision loss can be slowed or even averted. Currently, DR grading is primarily a manual process that is time-consuming and requires trained clinicians to evaluate digital retinal fundus images (RFIs). Time is of the essence here because delayed results can lead to delayed treatment, or even lost follow-up communication. Therefore, the need for an automatic DR grading and screening method has long been recognized. 


%
%

Following the theoretical evolution of deep neural networks (DNN), DNN-based automatic medical imagery analysis has become vastly popular in recent years~\cite{gulshan2016development,mansour2018deep}, as an aid to human experts. With the fast development of DNN-based image analysis and recognition techniques, less human intervention is required and, over time, the recognition systems can become fully automated. 
DR grading based on retinal fundus image analysis is one such popular domain where automatic DNN-based systems are deployed~\cite{li2019canet,He2020CABNet}. For example, Kaggle introduced a DR detection competition where the DR severity is labeled as one of the five levels: `0' for no DR, `1' for mild DR, `2' for moderate DR, `3' for severe DR, and `4' for proliferative DR~\cite{kaggledr}. This competition has drawn significant attention and supported a series of DR grading works~\cite{li2019canet}. Moreover, as indicated in \cite{gulshan2016development}, DNN has been used as an assessment of diabetic retinopathy so that patients can bypass the need to wait weeks for an ophthalmologist to review the retinal images.

\begin{figure}[t]
\centering
\includegraphics[width=1\linewidth]{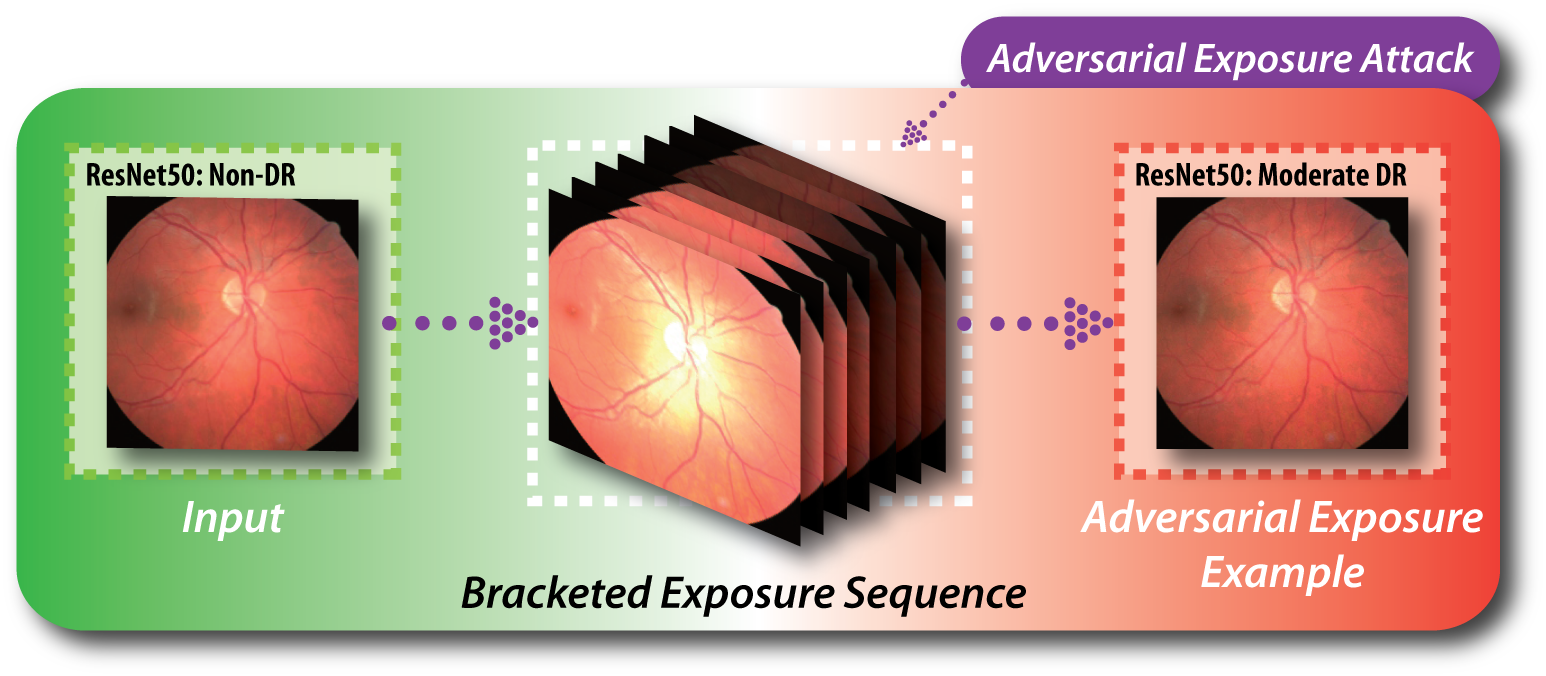}
\caption{A normal retinal fundus image (left) is correctly classified as `non-DR' by a pre-trained ResNet50. After applying our adversarial exposure attack, the corresponding adversarial example is mis-predicted as `moderate DR'.}
\label{fig:1}
\end{figure}


Despite its clear advantage in efficiency, DNN-based DR grading does have some caveats, especially when faced with fundus images that exhibit degradations~\cite{Fu2019EyeQ}. 
Low-quality fundus images can lead to higher uncertainty in clinical observation and the risk of misdiagnosis. One major cause of this is uneven light variation. 
As a human-centered study of a DNN-based DR detection system deployed in clinics discussed in \cite{beede2020human}, the bad lighting situation results in low-quality fundus images and disables the deployed DR grading system.
%
%
Note that, camera exposure is an inherent factor of an imaging system and can lead to the light variation in an image directly \cite{ray2000camera}. 
Hence, exposure variation is also an important but unexplored factor affecting DNN-based DR grading significantly, and it is important to develop new methods to comprehensively evaluate and enhance the generalization of deep learning models to the real-world exposure variation. 
One potential solution for investigating the effects of the exposure on DNNs is constructing a large scale DR dataset where each clean image has different counterparts caused by different exposures. However, it is hard to control the exposure and cover all possible situations in the real world. Besides, it would take significant effort to collect and annotate the correct grades.
%

Considering the difference between general and clinical scenarios, the impact of exposure is more common in RFI collection. In darkroom, no light leakage is allowed. Due to the strict and cumbersome shooting conditions that require patient cooperation, once the lesion is correctly displayed (in human's perception) after shooting, doctors will not deliberately optimize or reshoot the images. 
In contrast, the impact of exposure on general images is relatively small due to adequate light sources and many optimization methods. Furthermore, with the update of shooting equipment, the shooting of general images has become very easy, requiring only pressing the shutter button. This makes the cost of reshooting extremely low, increasing the probability of removing images affected by exposure.
Finally, the special and sensitive characteristics of medical data make it more challenging to create a large-scale DR dataset with enough exposure variations. Consequently, the demand for alternative solutions in the medical field is significantly higher than in the general image field.

In conclusion, the aim of our study is to evaluate the potential threat of camera exposure on AI models. However, image exposure is influenced by various factors, such as image rendering, shooting environment, and lighting conditions. Natural image exposure cannot cover all possible situations.
According to literature on exposure \cite{cai2018learning}, it has been noted that natural image exposures of scenes or images are mainly modified by the Exposure Value (EV) setting. However, simply changing the EV setting cannot effectively mislead a classification model and provide significant distortion. Another approach to generating exposure involves independently changing the exposure value of each pixel. But, this approach generates too much noise pattern. Fig.~\ref{fig:2} (b) illustrates this phenomenon.
Considering the large number of exposure situations, it is impractical to randomly adjust and generate natural image exposure to exhaust all possible exposures. Therefore, using natural image exposure to reveal the weakness of the model is not a suitable and efficient method.

To address the challenges and the clinical need, we implement the robustness verification of the DNN-based DR grading models through studying the influence of camera exposure from the viewpoint of adversarial attack. Specifically, we set out to reveal the vulnerabilities of DNN-based DR grading by carefully tuning the image exposure to mislead the DNNs and name the new task \textit{adversarial exposure attack}. Moreover, the generated adversarial examples can help enhance the DNN models via the adversarial training technique.

Our main contributions are summarized as follows: 
Firstly, we follow existing additive-perturbation-based adversarial attacks and implement a straightforward method, \revised{which is} \textit{multiplicative-perturbation-based exposure attack}, showing the challenges of this new task: \textit{it is difficult to generate natural adversarial exposure images with high attack transferability across DNNs}.
Secondly, to address the challenges, we propose the \textit{adversarial bracketed exposure fusion}, regarding the exposure attack as an element-wise bracketed exposure fusion problem in the Laplacian-pyramid space to increase the image naturalness of adversarial outputs. 
Moreover, we further propose the \textit{convolutional bracketed exposure fusion} in which the element-wise convolution takes the place of multiplication to enhance the attack transferability.
Finally, with the experiments conducted on a real-world public DR dataset, we show that our adversarial examples can achieve high image quality with a high attack transferability and help enhance the robustness of DNNs to exposure variations.

\revisedd{
The paper is organized as follows: Sec.~\ref{sec:related} introduces related works on DR grading and adversarial attacks. Sec.~\ref{sec:method} presents our proposed method. Sec.~\ref{sec:exp} describes the experimental setup and demonstrates the advantages of our algorithm based on the results. Finally, Sec.~\ref{sec:concl} concludes the paper.
}

\section{Related Work}\label{sec:related}

\subsection{Diabetic Retinopathy Grading}
\label{subsec:DRgrading}
Traditional automatic DR grading methods are based on handcrafted features.
They utilize retinopathic manifestations, including exudates, hemorrhages, and microaneurysms, as well as normal retina components such as blood vessels and optic discs.
Silberman \etal~\cite{silberman2010case} extracted scale-invariant feature transform (SIFT) features from input images. Then, they trained a support vector machine (SVM) classifier to recognize the exudates in the retinal image and predict different stages of DR depending on the result.
Akram \etal~\cite{akram2014detection} used filter banks and a hybrid classifier, which consists of an m-Mediods based model and Gaussian mixture model, to achieve lesion detection for grading DR.
%
%
Kumar \etal~\cite{kumar2017kernel} extended the multivariate generalized-Gaussian distribution to a reproducing kernel Hilbert space to generate a kernel generalized-Gaussian mixture model (KGGMM) for robust statistical learning.

In the past few years, many researchers have addressed this problem with the help of DNNs. 
\revised{
Yang \etal~\cite{yang2017lesion} utilized the annotations of location information, for example, microaneurysms, to design a two-stage DNN network for joint lesion detection and DR grading.
}
Gargeya \etal~\cite{gargeya2017automated} identified DR severity with a DNN classification model. 
\revised{
More recently, models that leverage the advantages of DNNs for assisting in the diagnosis of DR have been consistently explored \cite{khanna2023deep, mohanty2023using, parthiban2023diabetic,gadekallu2023deep}.
}
Besides, some papers~\cite{gulshan2016development,krause2018grader,li2019canet} have tried to explore and utilize the internal relationship between DR and diabetic macular edema (DME) to improve the performance in grading both diseases.
Gulshan \etal~\cite{gulshan2016development} built a DNN model based on the Inception-v3 architecture for grading DR and DME. 
Later, Krause \etal~\cite{krause2018grader} provided better results with an Inception-v4 architecture.
Afterwards, CANet~\cite{li2019canet} integrated a disease-specific attention module as well as a disease-dependent attention module in a unified network to further improve the overall performance on grading DR and DME. Although they have achieved great progress, existing works do not consider the influence of camera exposure, which frequently occurs in the diagnosis process.
\begin{figure}[t!]
\centering
\includegraphics[width=1\linewidth]{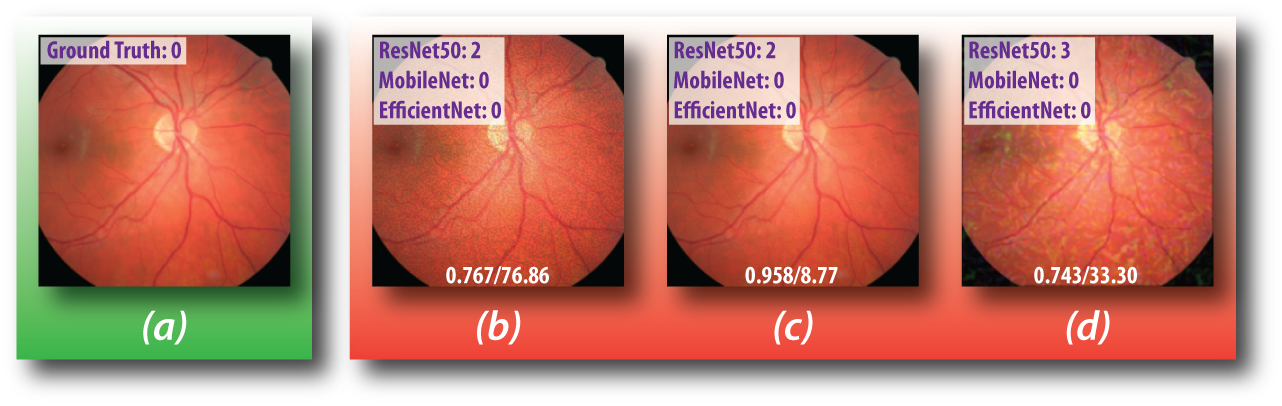}
\caption{Recognition visualization results of an example input (a) and adversarial examples (b-d) generated by different versions of our method discussed in Sec.~\ref{sec:method}. (b) is produced by Eq.~\eqref{eq:expmodel}. (c) is produced by bracketed exposure fusion (BEF). (d) is produced by convolutional bracketed exposure fusion (CBEF). The ground truth label is listed in the top-left of the input image. For each adversarial example, the predictions of three models, \revised{which are} ResNet50, MobileNet and EfficientNet, are listed in the top-left. 
The two values on the bottom refer to the image quality assessment values, which are \revised{structural similarity (SSIM) \cite{wang2004image} and blind/referenceless image spatial quality evaluator (BRISQUE) \cite{mittal2012no}}.}
\label{fig:2}
\end{figure}

\subsection{Adversarial Attacks}
\label{DR_subsec_advattack}
DNN techniques have facilitated numerous applications of artificial intelligent, including image classification~\cite{he2016deep},  detection~\cite{he2017mask}, segmentation~\cite{chen2021semantically}, salience object detection~\cite{xiao2018deep}, as well as medical tasks~\cite{yang2017lesion,gargeya2017automated,li2019canet}. 
%
%
However, recent studies have indicated that even imperceptible perturbations, called adversarial examples~\cite{szegedy2013intriguing}, can totally fool a well-trained DNN model generating wrong prediction result.
Generally, there are two kinds of adversarial attacks: whitebox attacks and blackbox attacks. 
Under a whitebox attack, the attack method has full access to the DNN model. Szegedy \etal~\cite{szegedy2013intriguing} first addressed the generation of adversarial examples as an optimization problem. Goodfellow \etal then proposed a one-step method to efficiently produce adversarial examples, which they named the fast gradient sign method (FGSM). 
Later, Kurakin \etal~\cite{kurakin2016adversarial} utilized iterative optimization to improve the performance. 
Soon after, Dong \etal~\cite{dong2018boosting} further upgraded this method by applying a momentum term. 
Afterwards, Dong \etal~\cite{dong2019evading} explored how to enhance the transferability of adversarial examples. 
\revised{
Furthermore, Lin \etal~\cite{LinS00H20} utilized Nesterov accelerated gradient to achieve higher transferability.
Then, Wang \etal~\cite{wang2021enhancing} also reinforced the transferability via variance tuning.
}
Other kinds of whitebox attacks, such as DeepFool~\cite{moosavi2016deepfool}, sacrifice time complexity to generate tiny perturbations in a simple way. 
Papernot \etal~\cite{papernot2016limitations} achieved an adversarial attack by restricting the $\ell_0$ norm, which perturbs only a few pixels in the image. 
Su \etal~\cite{su2019one} proposed a method that conducts an adversarial attack by modifying a single pixel.
Carlini \etal~\cite{carlini2017towards} produced highly imperceptible perturbations by optimizing crafted object functions under different norms.
Similar to this work, Zhang \etal~\cite{zhang2020walking} introduced a new attack called Boundary Projection (BP), which explores a speed-distortion trade-off strategy and adapts the distortion of adversarial examples in each step.
Moreover, Francesco Croce and Matthias Hein~\cite{croce2019sparse} proposed a highly sparse adversarial attack aimed at minimizing the $L_0$ distance to the original image while making the distortion imperceptible.
To provide robust adversarial training, Aleksander Madry \etal~\cite{MadryMSTV18} proposed a powerful attack named Projected Gradient Descent (PGD), considering the strongest first-order attack method.
\revised{
Schwinn \etal~\cite{schwinn2023exploring} introduced a new loss function designed for adversarial attacks, which provides consistent improvement comparing with old loss function.
}
Rather than using an additive attack, Guo \etal~\cite{guo2020abba} recently introduced an innovative way to attack the input by a blurring operation.
\revised{
Additionally, physical adversarial attacks have been shown to pose severe threats to DNN-based models. Wei \etal~\cite{wei2023unified} introduced a unified adversarial patch for conducting cross-modal physical attacks. Utilizing a novel boundary-limited shape optimization, they designed compact and smooth shapes that are easily applicable in the physical environment.
Wang \etal~\cite{wang2023rfla} positioned a colored transparent plastic sheet and a paper cutout of a specific shape in front of a mirror to perform a new Reflected Light Attack (RFLA) in digital and physical world.
The investigation of adversarial attacks across various input formats has also got attention in recent years. Kim \etal~\cite{kim2023breaking} introduced the ``Breaking Temporal Consistency'' method to generate universal adversarial perturbations, marking the first incorporation of temporal information into video attacks.
}
In a blackbox attack, the attacker has no prior information of the target DNN model.
Under this setting, Chen \etal~\cite{chen2017zoo} produced adversarial examples by estimating the gradients of the target model.
\revised{
Gubri \etal~\cite{gubri2022lgv} presented a novel approach named Large Geometric Vicinity aimed at improving the transferability of black-box adversarial attacks.
}
\revised{
Pomponi \etal~\cite{9892966} introduced a new method that successfully attacks a large proportion of samples by rearranging a bit number of pixels in the image.
}
Other works~\cite{baluja2017adversarial, hayes2017machine} trained an attacker neural network to achieve blackbox attacks. 
Besides, Guo \etal~\cite{guo2019simple} proposed a simple way of constructing the adversarial perturbation in the discrete cosine transform (DCT) space and reached similar blackbox attack capability, but, with fewer search steps. Wang \etal~\cite{wang2020amora} achieved a blackbox attack through facial morphing instead of additive noise.
More recently, Yoo \etal~\cite{yoo2020outcomes} built binary classifier models to detect diabetic retinopathy or diabetic macular edema. They found that even simple FGSM can fool the classifiers.
In addition to the above attack methods,  Jan \etal~\cite{jan2019connecting} studied how to make the generated adversarial examples robust to physical camera transformations. To this end, they trained a conditional GAN to simulate a physical image of a clean image and then added adversarial noise to the `physical’ counterpart.

Note that existing adversarial attacks mainly focus on additive noise perturbations, which are hardly found in our daily life and thus do not help analyze the effects of real-world degradation, \revised{for example}, camera exposure. This work proposes an entirely new exposure-based adversarial attack simulating the real bracketed exposure fusion of photographs, which can help reveal the potential threats to DNN-based automated DR diagnosis. Moreover, to address the challenges stemming from naturalness requirements, we propose the adversarial bracketed exposure fusion, regarding the exposure attack as an element-wise bracketed exposure fusion problem in the Laplacian-pyramid space.

\section{Methodology}\label{sec:method}

\begin{figure*}[ht!]
\centering
\includegraphics[width=.9\linewidth]{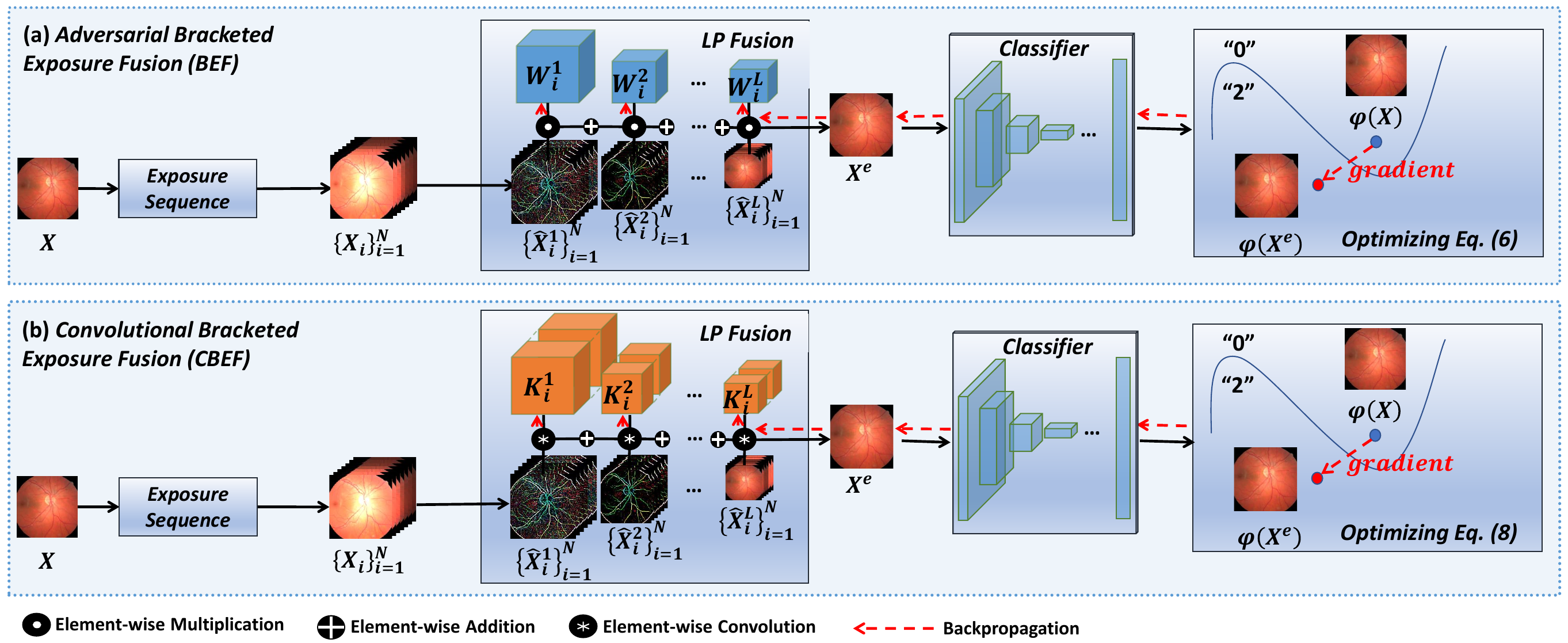}
\caption{Pipeline of our model for adversarial exposure attacks. \revised{(a) is based on adversarial BEF and (b) shows the workflow of CBEF}. Blue curve in the right block indicates the decision boundary between DR degrees '0' and '2'.
}
\label{fig:framework}
\end{figure*}

\subsection{Adversarial Exposure Attack on DR Imagery}
\label{DR_subsec_AE}
Given a DR image ($\mathbf{X}$), a pre-trained DNN ($\phi(\cdot)$) can estimate its grade by regarding it as a classification problem. Our goal is to investigate whether the DNN has vulnerability to common camera exposure. 
Following the general camera exposure process \cite{mertens2009exposure,yu2018deepexposure}, we can represent an exposure-degraded DR image ($\mathbf{X}^\text{e}$) as the multiplication between a clean image $\mathbf{X}$ and an exposure map $\mathbf{E}$ having the same size with $\mathbf{X}$, \revised{which is},
%
\begin{align}\label{eq:expmodel}
\mathbf{X}^\text{e} = \mathbf{X}\odot\mathbf{E},
\end{align}
%
where $\odot$ represents the element-wise multiplication. We can tune the exposure map in a pixel-wise way to produce different exposure images. Here, like existing additive-perturbation-based adversarial attacks~\cite{goodfellow2014explaining}, we change $\mathbf{E}$ according to the classification loss and achieve a multiplicative-perturbation-based adversarial attack against a pre-trained DNN ($\phi(\cdot)$) by optimizing the following objective function:
%
\begin{align}\label{eq:obj}
\argmax_{\mathbf{E}}{J(\phi(\mathbf{X}\odot\mathbf{E}),y)},
\text{~subject~to~} \|\mathbf{E}-\mathbf{1}\|_\text{p}<\epsilon,
\end{align}
%
where $J(\cdot)$ represents the image classification loss function \revised{(cross-entropy function)}, and $y$ is the ground truth grade of $\mathbf{X}$. $\epsilon$ controls the perturbation degree. Intuitively, the exposure map $\mathbf{E}$ is modified in a pixel-wise way to maximize the objective function under the constraint of $\epsilon$. 
A large value of the objective function means the DNN $\phi(\cdot)$ tends to mis-grade the example $\mathbf{X}^\text{e}$.
A small $\epsilon$ lets the $\mathbf{E}$ approximate to one, making $\mathbf{X}^\text{e}$ similar to $\mathbf{X}$. As a result, the modification would not be easily perceived.

Although the above method is simple and easy to implement, we argue that it is not easy to achieve the desired exposure attack for several reasons: \ding{182} Camera exposure usually leads to spatially smooth variation across the DR images. However, the above method generates noise patterns for the images (\revised{for example}, Fig.~\ref{fig:2} (b)) that are easily perceived and different from the natural camera exposure, making them less meaningful for the exposure-robustness analysis of DNNs. 
To address this problem, we propose the \textit{adversarial bracketed exposure fusion} based attack in Sec.~\ref{subsec:advexpfusion}, which can generate smooth local variation with the Laplacian pyramid representation and adversarially tuned fusion weight maps.
\ding{183} From the viewpoint of adversarial attacks, the above method cannot achieve high transferability when the adversarial exposure example is generated for one DNN and used to attack another. 
Intuitively, the adversarial exposure example fooling more DNNs (\revised{which means} high transferability attack) are more meaningful for investigating the exposure-robustness of DNNs.    
To this end, we further extend the multiplicative operation to the convolution operation in Sec.~\ref{subsec:ConvExposureFusion}, enabling us to achieve high transferability attack with spatially smooth intensity variation.

\subsection{Adversarial Bracketed Exposure Fusion}\label{subsec:advexpfusion}
%
Inspired by this technique, we regard the adversarial exposure attack as a bracketed exposure fusion problem \cite{mertens2009exposure}, where multiple bracketed exposure images are first generated and fused within the Laplacian-pyramid space. 
Intuitively, the fusion on Laplacian-pyramid space benefits the smooth requirements, and the adversarially tuned fusion weight maps guarantee the capability of fooling DNNs. 

\subsubsection{Bracketed Exposure Sequence Generation} 
As shown in Fig.~\ref{fig:framework}~(a), given a DR image $\mathbf{X}$, we first generate an exposure sequence $\{\mathbf{X}_i\}_{i=1}^{N}$ by multiplying the raw image with different exposure scalars, \revised{which is},
%
\begin{align}\label{eq:bracketed}
\mathbf{X}_i=\mathbf{X}\cdot2^{e_i},
\end{align}
%
where $e_i$ denotes the exposure shifting value, which we set within the range of $[-\lambda,\lambda]$ with $\lambda>0$. 
The exponential function based on $2$ is used for converting an exposure shifting $e_i$ to the intensity transmission \cite{ray2000camera}.
Intuitively, when $e_i>0$, $\mathbf{X}$ will be brighter, otherwise it will be darker. 
According to Eq.~\eqref{eq:bracketed}, we can produce $N$ images with different exposures. Then, we aim to fuse these images to generate the final adversarial example which looks natural while being able to fool DNNs.

\subsubsection{Bracketed Exposure Fusion} 
\label{DR_subsec_BEF}
%
%
As depicted in Fig.~\ref{fig:2} (b), while the adversarial examples effectively mislead the ResNet50 classifier, their image quality assessment values are low, with $0.767$ for SSIM (higher is better) and $76.86$ for BRISQUE (lower is better).
Motivated by \cite{mertens2009exposure}, we integrate Laplacian pyramid fusion into our attack strategy to suppress the noise-like patterns. 
Concretely, the Laplacian pyramid is a multi-scale signal representation that enforces smoothness across local pixels \cite{mertens2009exposure}, thus it is suitable for conducting adversarial exposure attack. 
In brief, given two images, BEF first represents them in the Laplacian pyramid space and obtains band-pass filtered images at each scale ~\cite{burt1983multiresolution}. Image fusion is then processed at each scale, respectively. After that, the images in all levels are composed to generate the final seamless fusion result.
Benefited by the multi-scale fusion, BEF can help to seamlessly blend those bracketed exposure images by combing them into different resolutions.
Fig.~\ref{fig:2} (c) demonstrates the adversarial outcome achieved with Laplacian pyramid fusion for the same input. It achieved better image quality under the condition of successful attack, with $0.958$ for SSIM and $8.77$ for BRISQUE.

Specifically to our case, we decompose each bracketed exposure image ($\mathbf{X}_i$) to the Laplacian pyramid (LP) representation \cite{mertens2009exposure} and assign each bracketed exposure image weight maps $\{\mathbf{W}_i^l\}$ in the Laplacian-pyramid space. 
The process is represented as $\mathcal{L}(\mathbf{X}_i)=\{\hat{\mathbf{X}}_i^l\}_{l=1}^L$ where $\hat{\mathbf{X}}_i^l$ is the $l$-th level decomposition of $\mathbf{X}_i$ and $\mathbf{W}_i^l$ refers to the weight map of $\hat{\mathbf{X}}_i^l$. Intuitively, each layer, $\hat{\mathbf{X}}_i^l$, serves as a band-pass filtering result of the exposure sequence ${\mathbf{X}}_i$.
Then, we can fuse the bracketed exposure images with their weight maps at each level
%
\begin{align}\label{eq:bracketed_fusion}
\hat{\mathbf{X}}^{\text{e},l}=\sum_{i=1}^{N} \mathbf{W}_i^l\odot\hat{\mathbf{X}}_i^l,
\text{~subject~to~~} \forall l,\sum_{i=1}^{N} \sum_{p} \mathbf{W}_{i,p}^l=\mathbf{1},
\end{align}
%
where $\mathbf{W}_{i,p}^l$ is $p$-th scalar element of $\mathbf{W}_i^l$. The constraint term means $\{\mathbf{W}_i^l\}$ should be normalized at each level. Then, the fusion results of all levels, \revised{} $\{\hat{\mathbf{X}}^{\text{e},l}\}_{l=1}^{L}$, are used to reconstruct the final result
%
\begin{align}\label{eq:bracketed_reconstruction}
\mathbf{X}^\text{e}=\mathcal{L}^{-1}(\{\hat{\mathbf{X}}^{\text{e},l}\}_{l=1}^{L})=\mathcal{L}^{-1} \Big( \{\sum_{i=1}^{N} \mathbf{W}_i^l\odot\hat{\mathbf{X}}_i^l\}_{l=1}^{L} \Big),
\end{align}
%
where $\mathcal{L}^{-1}(\cdot)$ denotes the inverse LP decomposition. The LP fusion module in Fig.~\ref{fig:framework}~(a) shows the process.
%
We do not directly adjust the exposure map in Eq.~\eqref{eq:expmodel}. Instead, we modify the weight maps in a layer-wise and pixel-wise manner, and use Eq.~\eqref{eq:bracketed_reconstruction} to define the \textit{adversarial bracketed exposure fusion} based attack. This approach enables us to tune all weight maps such that the resulting fused DR image can deceive DNNs.
%
\begin{align}\label{eq:bracketed_obj}
\argmax_{\{\mathbf{W}_i^l\}} & {~J \Bigg( \phi \bigg( \mathcal{L}^{-1} \Big( \{\sum_{i=1}^{N} \mathbf{W}_i^l\odot\hat{\mathbf{X}}_i^l\}_{l=1}^{L} \Big) \bigg), y \Bigg) },\\ \nonumber
& \text{~subject~to~~} \forall l, \sum_{i=1}^{N} \sum_{p} \mathbf{W}_{i,p}^l=\mathbf{1}.
\end{align}
%

As shown in Fig.~\ref{fig:2} (c), the proposed BEF attack can generate smooth and natural adversarial examples. 
Nevertheless, according to our evaluation, such an attack is still unable to achieve high transferability across different models. One possible reason is that the linear fusion via element-wise weight maps cannot efficiently represent complex perturbation patterns that could fool DNNs.

\subsection{Convolutional Bracketed Exposure Fusion}\label{subsec:ConvExposureFusion}
To address the low transferability of the \textit{adversarial bracketed exposure fusion} based attack, we extend the element-wise linear fusion to a convolutional strategy where more parameters can be tuned. We re-formulate Eq.~\eqref{eq:bracketed_reconstruction} as
%
\begin{align}\label{eq:conv_bracketed_reconstruction}
\mathbf{X}^\text{e}=\mathcal{L}^{-1}(\{\hat{\mathbf{X}}^{\text{e},l}\}_{l=1}^{L})=\mathcal{L}^{-1} \Big( \{\sum_{i=1}^{N} \mathbf{K}_i^l\circledast\hat{\mathbf{X}}_i^l\}_{l=1}^{L} \Big),
\end{align}
%
where $\circledast$ denotes the element-wise convolution. 
In element-wise convolution, each pixel of $\hat{\mathbf{X}}_i^l$ is processed by a specific kernel that is exclusively stored in $\mathbf{K}_i^l$. For example, if $\hat{\mathbf{X}}_i^l\in\mathds{R}^{H\times W}$, then the kernel $\mathbf{K}_i^l$ should have a size of $\mathds{R}^{H\times W\times K^2}$. This means that the $p$-th pixel of $\hat{\mathbf{X}}_{i,p}^l$ and its $K\times K$ neighboring pixels, termed as $N_K(\hat{\mathbf{X}}_{i,p}^l)$, are combined linearly using the weight map $\mathbf{K}_{i,p}^l\in\mathds{R}^{K\times K}$, which is the reshaped version of the $p$-th element of $\mathbf{K}_i^l$. Here, $K$ represents the kernel size. Therefore, the new $p$-th pixel is calculated by:
${\hat{\mathbf{X}}'^l_{i,p}} = N_K(\hat{\mathbf{X}}_{i,p}^l) * \mathbf{K}_{i,p}^l$.
With Eq.~\eqref{eq:conv_bracketed_reconstruction}, we can reformulate Eq.~\eqref{eq:bracketed_obj} as 
%
\begin{align}\label{eq:conv_bracketed_obj}
\argmax_{\{\mathbf{K}_i^l\}} & {~J \Bigg( \phi \bigg( \mathcal{L}^{-1} \Big( \{\sum_{i=1}^{N} \mathbf{K}_i^l\circledast\hat{\mathbf{X}}_i^l\}_{l=1}^{L} \Big) \bigg),y \Bigg) },\\ \nonumber
& \text{~subject~to~~} \forall l, \sum_{i=1}^{N} \sum_{p,q}\mathbf{K}_{i,p,q}^l=\mathbf{1},
\end{align}
%
where $\mathbf{K}_{i,p,q}^l$ is a scalar and denotes the $q$-th element of $\mathbf{K}_{i,p}^l$.

\subsection{Optimization and Attack Algorithm}
\subsubsection{BEF}

Following existing adversarial attack methods \cite{szegedy2013intriguing, kurakin2016adversarial, dong2018boosting}, we solve Eq.~\eqref{eq:bracketed_obj} via the widely used sign gradient descent.
Specifically, for each weight map ($\mathbf{W}_{i}^l$ in Eq.~\eqref{eq:bracketed_obj}), we optimize it via
%
\begin{align}\label{eq:bracketed_obj_opt}
\mathbf{W}_{i,t}^l=\mathbf{W}_{i,t-1}^l+\alpha_{w}~\text{sign}(\nabla_{\mathbf{W}_{i,t-1}^l}J),
\end{align}
%
where $\text{sign}(\cdot)$ denotes the sign function and $\alpha$ is the optimizing step size. 
The attacking process can be simply summarized as follows:
\textit{First}, we initialize the weight map $\mathbf{W}_i^l$ as an identity counterpart letting $\hat{\mathbf{X}}_i^l=\mathbf{W}_i^l\odot\hat{\mathbf{X}}_i^l$.
\textit{Second}, we calculate $\mathbf{X}^\text{e}$ via Eq.~\eqref{eq:bracketed_reconstruction}.
Then, we calculate the loss via $J(\cdot)$ in Eq.~\eqref{eq:bracketed_obj}, perform back-propagation, and update $\mathbf{W}_i^l$ via Eq.~\eqref{eq:bracketed_obj_opt}. After that, we go back to the second step for further optimization until the maximum iteration number is reached.
The red dash line in Fig.~\ref{fig:framework} illustrates the back-propagation of the gradient.

\subsubsection{CBEF}
\label{DR_subsec:CBEF}
As shown in Fig.~\ref{fig:framework}~(b), generally, the optimization of CBEF is similar to that of BEF. 
CBEF extends the pixel-wise weigh maps $\mathbf{W}_i$ in BEF to kernel maps $\mathbf{K}_i$. Consequently, the CBEF-based attack is to optimize the kernels $\{\mathbf{K}_i^l\}$ instead of $\{\mathbf{W}_i^l\}$ and we have 
%
\begin{align}\label{eq:conv_bracketed_obj_opt}
\mathbf{K}_{i,t}^l=\mathbf{K}_{i,t-1}^l+\alpha_k~\text{sign}(\nabla_{\mathbf{K}_{i,t-1}^l}J),
\end{align}
%

\subsection{PGD embedded BEF and CBEF}
\label{DR_subsec:PGD_BEF_CBEF}
To further validate the potential of our method, we implement the optimizations of BEF and CBEF (\revised{} Eq.~\eqref{eq:bracketed_obj_opt} and Eq.~\eqref{eq:conv_bracketed_obj_opt}) via the PGD~\cite{MadryMSTV18} attack, which are denoted as \BEFPGD{} and \CBEFPGD{}, respectively.
Specifically, we adopt the PGD attack \cite{MadryMSTV18} to randomly initialize the weight matrix $\mathbf{W}$ and compute the loss using the image classification loss function $J(\cdot)$ as shown in Eq.~\eqref{eq:bracketed_obj}. The resulting \textbf{gradient} $\nabla J$ is then backpropagated from the output layer to the input layer $\mathbf{X}^e$  of the \textbf{Classifier} in Fig.~\ref{fig:framework}. Mathematically, this procedure can be expressed as:
%
\begin{align}\label{eq:bracketed_obj_opt_pgd_Xe}
\nabla^{pgd}_{\mathbf{X}^e} J= \nabla J ( \phi( \mathbf{X}^e ), y ),
\end{align}
%
As we have obtained the PGD gradient $\nabla^{pgd}_{\mathbf{X}^e_i} J$ for each $i$-th pixel in $\mathbf{X}^e$ using the PGD attack method, the next step is backpropagating $\nabla^{pgd}_{\mathbf{X}^e_i} J$ to the $\mathbf{W}$ layer, which generates $\nabla^{pgd}_{\mathbf{W}_{i,t-1}^l}J$. 
Therefore, we can rewrite the Eq.~\eqref{eq:bracketed_obj_opt} and update the $\mathbf{W}_{i,t}^l$ in \BEFPGD{} as follow:
%
\begin{align}\label{eq:bracketed_obj_opt_pgd}
\mathbf{W}_{i,t}^l=\mathbf{W}_{i,t-1}^l+\alpha_{w}~\text{sign}(\nabla^{pgd}_{\mathbf{W}_{i,t-1}^l}J).
\end{align}
%
Similarlly, the \CBEFPGD{} optimization function is:
%
\begin{align}\label{eq:conv_bracketed_obj_opt_pgd}
\mathbf{K}_{i,t}^l=\mathbf{K}_{i,t-1}^l+\alpha_{k}~\text{sign}(\nabla^{pgd}_{\mathbf{K}_{i,t-1}^l}J),
\end{align}
%
Moreover, we will discuss how PGD attack can benefit our method in aspect of the performance in next section.

In practice, we tune the attack step sizes $\alpha_w$ in Eq.~\eqref{eq:bracketed_obj_opt} and $\alpha_k$ Eq.~\eqref{eq:conv_bracketed_obj_opt} from $0.005$ to $0.1$ to generate adversarial outcomes with different image qualities and their corresponding attack success rate.
Moreover, we will discuss the influence of hyper-parameters including the level $L$ of the Laplacian-pyramid space in the adversarial bracketed exposure fusion as well as the kernel size $K$ in the convolutional bracketed exposure fusion in the experiment section.
Finally, according to Eq.~\eqref{eq:bracketed_obj} and Eq.~\eqref{eq:conv_bracketed_obj}, it should be noted that the number of parameters to be optimized for BEF and CBEF is $O(L \cdot N \cdot H \cdot W)$ and $O(L \cdot N \cdot H \cdot W \cdot K^2)$, respectively. In our experiments, conducted on a GTX 3090 GPU, the average run times for attacking a single image using our methods are as follows: BEF - 0.83s; CBEF - 1.82s; \BEFPGD{} - 0.73s; \CBEFPGD{} - 1.34s.

\section{Experiments}\label{sec:exp}
In this section, we demonstrate the capability of adversarial exposure attack as well as the transferability of our method for DR images. 
The transferability denotes the capability of using adversarial examples crafted from one DNN model to attack another one, which indicates the generalization of the adversarial examples and is of great importance for black-box attacking.
We first describe the experimental settings in Sec.~\ref{subsec:setup}. Then, we report the comparison results with the additive-perturbation-based baseline methods in Sec.~\ref{subsec:comparison}. Finally, we also explore the influence of each component of our method in Sec.~\ref{subsec:ablation}.

\subsection{\revised{Experimental Settings}}
\label{subsec:setup}

\begin{figure}[tp]
    \centering
    \subfigure{
    \includegraphics[width=0.7\linewidth]{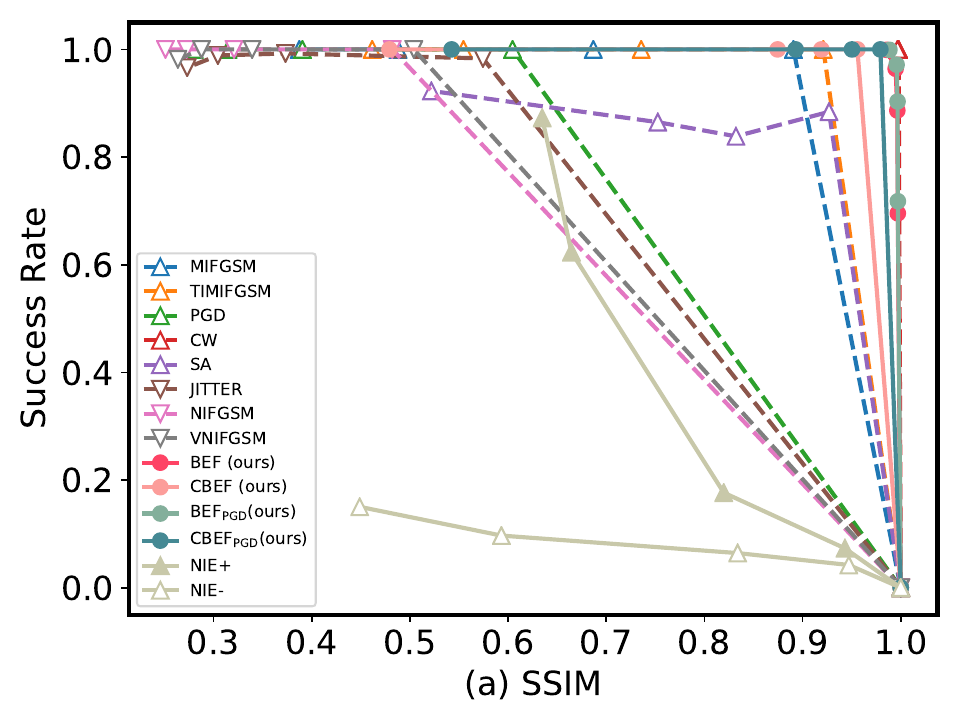}
    }
    \subfigure{
    \includegraphics[width=0.7\linewidth]{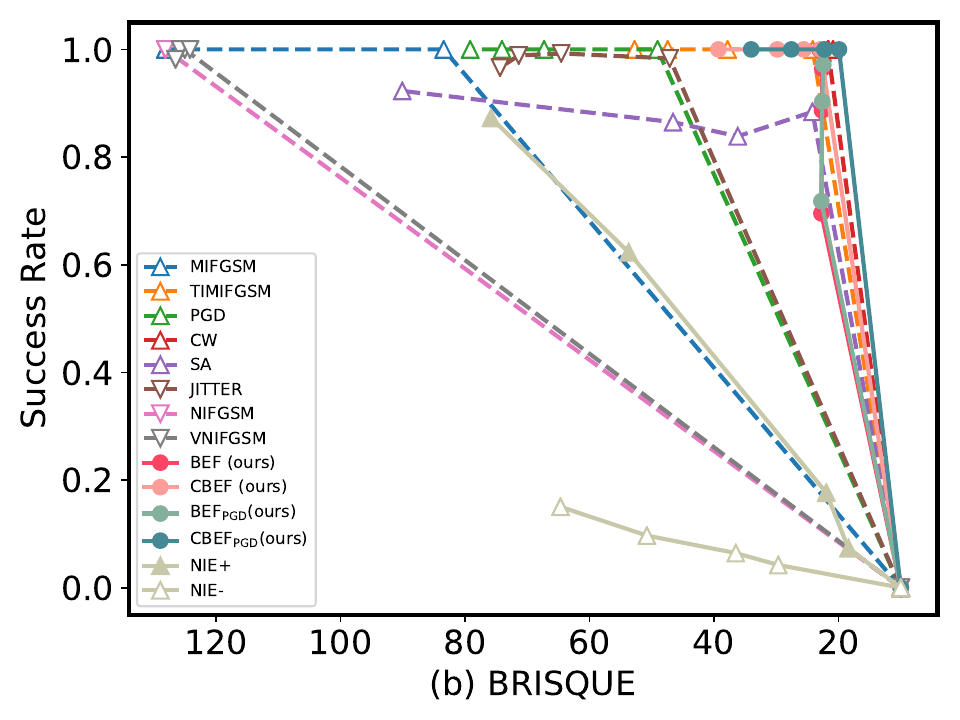}
    }
    \caption{Attack success rate along with SSIM and BRISQUE for adversarial examples crafted from ResNet50 by the baseline methods and our four attacks, \revised{which are} BEF, CBEF, \BEFPGD~and \CBEFPGD. \revisedd{(To reduce confusion caused by too many curves, we hid the less competitive baselines, FGSM, IFGSM, TIFGSM, and TIIFGSM.)} Our curves are generated by tuning the attack step sizes $\alpha_w$ in Eq.~\eqref{eq:bracketed_obj_opt} and $\alpha_k$ Eq.~\eqref{eq:conv_bracketed_obj_opt} from $0.005$ to $0.1$. For the additive-perturbation-based attacks, we tune the maximum perturbation ranges from $16$ to $64$ with the max intensity of $255$. For C\&W attack, we tune the weight $c$ ranging from 0.01 to 10. For SA attack, we tune the maximum perturbed pixels $k$ \revised{ranging from $1 \times 10^4$ to $5 \times 10^4$.} For NIE, we tune the EV setting $e_i$, in Eq.~\eqref{eq:bracketed}, from $-3$ to $3$ ("NIE+": $0$ to $3$. "NIE-": $0$ to $-3$).
    }
\label{fig:exp_whitebox}
\end{figure}

\subsubsection{Dataset}
\label{subsubsec:dataset}

We conduct all experiments on the EyePACS~2015 dataset \cite{EyePACS}, which contains 88,702 images and is one of the largest retinal image datasets and has been used in many recent DR related works~\cite{gargeya2017automated,zhou2019collaborative}. 
We use the entire training dataset (\revised{which contains} 35126 images) of EyePACS~2015 to train all DR grading deep models and randomly select 3,000 images from its testing dataset to perform attacks. Note that, the ratio of each grade level on the selected images is the same as the raw testing dataset.
The DR severity is graded from $0$-$4$\footnote{\scriptsize{Detailed information about the dataset can be found at \url{https://www.kaggle.com/c/diabetic-retinopathy-detection}}}. Specifically, $0, 1, 2, 3, 4$ refer to no DR, mild DR, moderate DR, severe DR, and proliferative DR, respectively.

\begin{figure*}[t]
    \centering
    \includegraphics[width=0.98\linewidth]{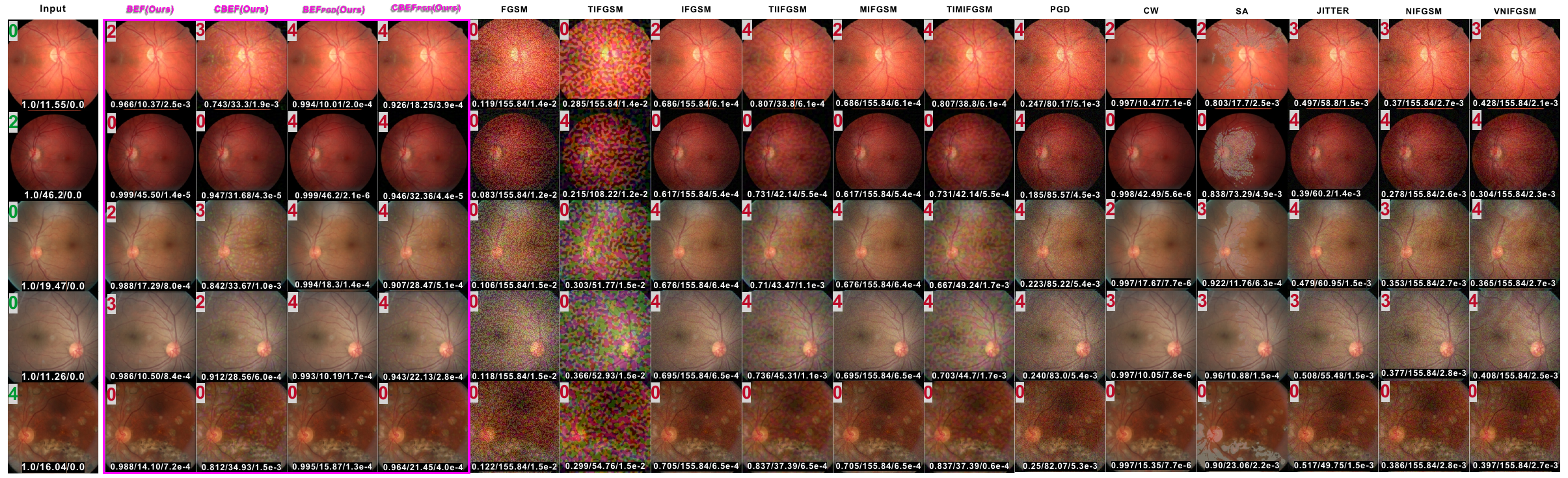}
    \caption{Visualization results of adversarial examples crafted for the ResNet50, using our methods, \revised{which are} BEF, CBEF, \BEFPGD{}, and \CBEFPGD{}, as well as baseline attacks. For each image, its DR grading result through ResNet50 is displayed on the top-left. The three numbers at the bottom are SSIM, BRISQUE and $L_2$ norm values ($a \mathrm{e} -b$ refers to \revised{$a \times 10^{-b}$}). The five inputs (1st column) are correctly classified to their ground truth DR grades, and the numbered labels can be used to gauge the classification of all the remaining attack algorithms (column $2$-$17$).}
\label{fig:via_baselines}
\end{figure*}

\subsubsection{Metrics}
\label{DR_subsec:metrics}
We select the attack success rate ($ASR$) to evaluate attack capability:

\begin{align}\label{eq:success_rate}
ASR = \cfrac{acc_{org} - acc_{adv}}{acc_{org}},
\end{align}
where $ASR$ indicates the success rate. $acc_{org}$ refers to the accuracy of original images without adversarial attack and $acc_{adv}$ stands for the accuracy of adversarial examples.
Besides, a desired adversarial exposure output should maintain natural for human perception since a retinal fundus image with severe perturbations would be easily disregarded by doctors or operating staff. To this end, we also introduce the image quality assessment (IQA) in our experiments for image quality (or naturalness) evaluation.
We choose SSIM \cite{wang2004image} as the image quality metric to evaluate the perception changes after attack through the local structure information~\cite{ram2017deepfuse,ma2015perceptual,cai2018learning}.
Moreover, we follow several exposure-related works \cite{zhang2018image,panetta2022deep,steffens2018deep,lee2020learning} that introduce BRISQUE \cite{mittal2012no} to further evaluate the naturalness of the output as it serves as a non-reference IQA metric based on the natural scene statistics.
%
Overall, we select attack success rate, SSIM, and BRISQUE for evaluating the performance of our method as well as all of the baselines.

\subsubsection{Models}
To evaluate the attack capability and transferability of our method against or across different neural networks, we introduce three widely-used DNNs, including ResNet50 \cite{he2016deep},  MobileNet \cite{howard2017mobilenets}, and EfficientNet \cite{tan2019efficientnet}. They are all pre-trained on the ImageNet dataset \cite{deng2009imagenet} and fine-tuned on the EyePACS~2015 dataset \cite{EyePACS}'s entire training dataset (35126 images).
%
%
Then, we evaluate the three models, \revised{which are} ResNet50, MobileNet, and EfficientNet, on our testing dataset (\revised{which contains} the 3000 images selected from EyePACS~2015's testing dataset). The evaluation results (\revised{} classification accuracy) are $80.0\%$, $82.4\%$, and $83.5\%$, respectively. 

\subsubsection{Baselines}
\label{DR_subsce:Baselines}
\revised{
We choose several additive-perturbation-based adversarial attacks as the baseline methods: FGSM \cite{goodfellow2014explaining},
iterative fast gradient sign method (IFGSM) \cite{kurakin2016adversarial}, momentum iterative fast gradient sign method (MIFGSM) \cite{dong2018boosting}, three translation-invariant (TI) enhanced versions \cite{dong2019evading} (denoted as TIFGSM, TIIFGSM, and TIMIFGSM), PGD attack \cite{MadryMSTV18}, 
JITTER attack \cite{schwinn2023exploring}, 
Nesterov Iterative Fast Gradient Sign Method (NIFGSM) \cite{LinS00H20}, and variance tuning NIFGSM (VNIFGSM) attack \cite{wang2021enhancing}. All methods use the infinity norm. 
}

%
\revised{
In addition, we have also employed additional types of challenging attacks to generate image perturbations, namely the C\&W~\cite{carlini2017towards,carlini2017adversarial} attack (with $L_2$ norm), and the projected gradient descent variant of sparse attack (SA)~\cite{croce2019sparse}.
}
Finally, we propose a basic baseline, called natural image exposure (NIE), by adjusting the EV setting of each image ($e_i$ in Eq.~\eqref{eq:bracketed}) from $-3$ to $3$ ("NIE+": $0$ to $3$, "NIE-": $0$ to $-3$). We utilize this baseline to illustrate the statement in Sec.~\ref{sec:intro}, \revised{which is} why we develop a specific attack instead of relying on natural image exposure to expose the threat posed by camera exposure.

\subsection{Comparison with Baseline Methods}
\label{subsec:comparison}

\begin{algorithm}[!th]

		\caption{\fontfamily{bch}\selectfont \footnotesize{Detailed experimental process of quantitative analysis on attack success rate}}
  \label{alg_quanti_whitebox}
		\KwIn{\\
  An DR images classification dataset containing images and labels $\{\mathbf{X}, \mathbf{Y}\}$; \\
  The pre-trained threatened ResNet50 model, $\{\phi(\cdot)\}$; \\
  Attack methods $\{atk_j(\cdot, \cdot, \alpha, \phi)\}_{j=1 \cdots J}$; \\
  Hyper-parameters set of each attack method $j$: $\{\alpha_{jk}\}_{k=1, \cdots, K}$;  \\
  Evaluation metrics ASR and IQA: $ASR(acc_{org}, acc_{adv})$, $IQA(\cdot)$}
		\KwOut{\\
  Curve of each attack method $atk_j$ against ResNet50 ($\phi(\cdot)$).\\
  }
  
  \For{$j=1\ \mathrm{to}\ J$}{
      \For{$k=1\ \mathrm{to}\ K$}{
        Generate the adversarial examples: $\textbf{X}_{adv}^{jk}$ = $atk_j(\textbf{X}, \textbf{Y}, \alpha_{jk}, \phi)$\\
        Calculate the accuracy of ResNet50 when predicts original images $\textbf{X}$: $acc_{org}$ \\
        Calculate the accuracy of ResNet50 when predicts adversarial examples $\textbf{X}_{adv}^{jk}$: $acc_{adv}^{jk}$ \\
        Calculate the ASR of $\textbf{X}_{atk}^{jk}$: $asr_{jk} = ASR(acc_{org}, acc_{adv}^{jk})$\\
        Calculate the IQA of $\textbf{X}_{adv}^{jk}$: $iqa_{jk}$ = $IQA(\textbf{X}_{adv}^{jk})$
      }
      Plot the curve of attack method $atk_j$ with ASR-IQA pairs $(asr_{jk}, iqa_{jk})_{k=1, \cdots, K}$\\ 
  }

\end{algorithm}%

\begin{algorithm}[!th]

{\caption{\fontfamily{bch}\selectfont \footnotesize{Detailed experimental process of quantitative analysis on transferability}}
\label{alg_quanti_transfer}
		\KwIn{\\
  An DR images classification dataset containing images and labels $\{\mathbf{X}, \mathbf{Y}\}$; \\
  The pre-trained threatened models (ResNet50, MobileNet and EfficientNet), $\{\phi_i(\cdot)\}_{i=1, \cdots, I}$; \\
  Attack methods $\{atk_j(\cdot, \cdot, \alpha, \phi)\}_{j=1 \cdots J}$; \\
  Pre-selected hyper-parameter of each attack method $j$: $\{\alpha_{j}\}$ based on the IQA results in Sec.~\ref{subsubsec:QuantitativeAna};  \\
  Evaluation metric of ASR: $ASR(acc_{org}, acc_{adv})$.}
		\KwOut{\\
  Table containing ASRs of the adversarial examples against each model $\phi_m(\cdot)$ (which are crafted from each threatened model $\phi_i(\cdot)$ by attack method $atk_j$).\\
  }

  \For{$i=1\ \mathrm{to}\ I$}{
      \For{$j=1\ \mathrm{to}\ J$}{
        Generate the adversarial examples: $\textbf{X}_{adv}^{ij}$ = $atk_j(\textbf{X}, \textbf{Y}, \alpha_{j}, \phi_i)$\\
        \For{$m=1\ \mathrm{to}\ M$}{
        Calculate the accuracy of each threatened model $\phi_m(\cdot)$ when predicts original images $\textbf{X}$: $acc_{org}^m$ \\
        Calculate the accuracy of each threatened model $\phi_m(\cdot)$ when predicts adversarial examples $\textbf{X}_{adv}^{ij}$: $acc_{adv}^{ijm}$ \\
        Calculate the ASR of $\textbf{X}_{adv}^{ijm}$: $asr_{ijm} = ASR(acc_{org}^m, acc_{adv}^{ijm})$\\
        }
      }
 }
Build table with $acc_{adv}^{ijm}$.\\  
}

\end{algorithm}%

\subsubsection{Quantitative Analysis on Attack Success Rate}
\label{subsubsec:QuantitativeAna}
In this part, we demonstrate the attack capability of our method by evaluating the success rate of adversarial examples crafted for ResNet50. We compare all methods based on the image quality of adversarial examples as well as the attack success rate for a fair comparison. Specifically, we tune parameters for different attacks to generate multiple success rate-SSIM/BRISQUE curves for clear visual comparison.
\revised{For example, we change the maximum perturbation values of additive-perturbation-based attacks ranging from $16$ to $64$ and the step size of our attack methods, $\alpha_w$ and $\alpha_k$ in Eq.~\eqref{eq:conv_bracketed_obj_opt}, from $0.005$ to $0.1$.
}
For C\&W attack, we tune the weight $c$ ranging from $0.01$ to $10$. 
For SA attack, we tune the maximum perturbed pixels $k$ ranging from \revised{$1 \times 10^4$ to $5 \times 10^4$}.
%
%
For NIE, we tune the EV setting $e_i$, in Eq.~\eqref{eq:bracketed}, from $-3$ to $3$.
\revised{
Experimental process is elaborated in Algorithm~\ref{alg_quanti_whitebox}.
}

We show the comparison results in Fig.~\ref{fig:exp_whitebox}, and observe that: 
\ding{182} 
Our methods (BEF, CBEF, \BEFPGD{}, and \CBEFPGD{}) can generate high-quality adversarial images (\revised{} higher SSIM scores or lower BRISQUE values) while achieving $100\%$ attack success rate, indicating the high capability in attacking the target models.
\ding{183} 
%
%
%
%
\revisedd{Compared to baseline methods, our BEF, CBEF, \BEFPGD{}, and \CBEFPGD{} achieve higher SSIM scores with similar attack success rates ($~100\%$). 
The C\&W attack is the only method that delivers comparable image quality, due to its optimization-based approach designed to minimize distortion. The key advantage of our methods lies in the Laplacian pyramid fusion, which effectively reduces noise. 
Moreover, our attacks outperform the baselines in BRISQUE scores, all around $20$, demonstrating that while our methods introduce perceptible changes, they maintain high image quality in terms of human perception.
}
\ding{184} 
\revisedd{The effectiveness of \BEFPGD{} over BEF is evident through improvements in both attack success rate and image quality. While the SSIM values of the first three data points in both methods are similar (ranging from $0.995$ to $0.998$), \BEFPGD{} achieves higher attack success rates. For example, the first point in \BEFPGD{} has a success rate of $71.5\%$ compared to BEF's $69.2\%$ with the same SSIM value of $0.998$. Both methods reach a final success rate of $99.9\%$, but \BEFPGD{} outperforms BEF with an SSIM of $0.988$ compared to BEF's $0.986$.
}
\ding{185} 
\revisedd{The NIE attack, which adjusts EV settings, performs poorly compared to other attacks. For instance, NIE reaches an $85.0\%$ success rate at an SSIM of $0.65$, but even baseline method like PGD achieves $100.0\%$ success at a similar SSIM value. This shows that altering EV settings alone cannot mislead the classification model and tends to cause significant distortion.}
%
\begin{table*}[ht]
\centering
\small
\caption{Adversarial comparison results on EyePACS~2015 dataset. We show the success rates ($\%$) of transfer \& whitebox adversarial attacks among three fine-tuned models: ResNet50, MobileNet, and EfficientNet, using additive attack baseline methods with a maximum perturbation of $32$, \revised{C\&W} attack with an adversarial loss weight $c$ of $1.0$, SA attack with a maximum number of perturbation pixels $k$ of \revised{$1 \times 10^4$} and four versions of our method (\revised{}BEF, CBEF, \BEFPGD{}, and \CBEFPGD{}). For each three columns, the whitebox attack results are shown in the last column. The first and second columns show the transfer attack results. We highlight the top three results with red, yellow, and green, respectively.} 

{
\resizebox{.95\linewidth}{!}{ 
\begin{tabular}{l|ccc|ccc|ccc}

\toprule

\rowcolor{tabgray}\multicolumn{1}{c|}{Crafted from} & \multicolumn{3}{c|}{ResNet50} & \multicolumn{3}{c|}{MobileNet} & \multicolumn{3}{c}{EfficientNet} \\

\cmidrule(r){1-1} \cmidrule(r){2-4} \cmidrule(r){5-7} \cmidrule(r){8-10}
\rowcolor{tabgray} Attacked model
&  MobileNet   &   EfficientNet &  ResNet50

&  ResNet50   &   EfficientNet &  MobileNet

&  ResNet50      &  MobileNet &   EfficientNet\\

\midrule

FGSM 
& 13.6  & 14.8    &11.0    
&7.7 &11.7 &10.7 
&7.5 &10.2 &13.9        \\

TIFGSM 
&13.7  &14.7    &11.0   
&7.7   &11.7 &10.5
&7.4 &10.3 &20.2        \\

IFGSM      
&14.3  &15.2    & \cellcolor{top1}100.0   
&\cellcolor{top2}47.4   &21.6 & \cellcolor{top1}100.0 
& \cellcolor{top2}54.7 &21.9 & \cellcolor{top1}100.0        \\

TIIFGSM      
&\cellcolor{top1}62.9  &18.9    &\cellcolor{top1}100.0   
&15.8   &14.1 &\cellcolor{top1}100.0 
&14.4 &32.2 &\cellcolor{top1}100.0        \\

MIFGSM      
& 14.0  & 15.0    & \cellcolor{top1}100.0   
&18.3   & 21.5 & \cellcolor{top1}100.0 
&26.6 & 16.8 & \cellcolor{top1}100.0       \\

TIMIFGSM      
&\cellcolor{top2}57.8  &20.9    &\cellcolor{top1}100.0   
&13.6   &15.5 &\cellcolor{top1}100 
&12.6 &24.4 &\cellcolor{top1}100.0        \\

PGD  
&14.2 &15.6  &\cellcolor{top1}100.0     
&10.4 &14.7 &\cellcolor{top1}100.0     
&15.4 &11.3 &\cellcolor{top1}100.0   \\

C\&W 
&22.6	&\cellcolor{top3}31.4	&\cellcolor{top1}100.0	
&21.7	&54.4	&\cellcolor{top1}100.0	
&35.0	&\cellcolor{top2}66.5	&\cellcolor{top1}100.0   \\ 

SA 
&39.7	&\cellcolor{top2}42.0	&88.4    
&\cellcolor{top3}38.3	&43.2	&84.3		
&37.9	&40.3	&91.1   \\

JITTER
&13.9	&15.1	&99.2 
&8.5	&20.7	&91.9		
&13.7	&12.5	&98.0   \\

NIFGSM
&14.4	&15.1	&\cellcolor{top1}100.0    
&7.7	&23.3	&99.9		
&7.6	&16.6	&\cellcolor{top1}100.0   \\

VNIFGSM
&30.9	&16.1	&\cellcolor{top1}100.0    
&7.9	&\cellcolor{top1}81.5	&\cellcolor{top1}100.0		
&7.7	&38.3	&\cellcolor{top1}100.0   \\


BEF (ours)      
&11.3  &13.4 &89.2   
&8.8   &14.8 &95.4 
&9.5 &17.5 &84.2        \\

CBEF (ours)      
&27.6  &29.7 &\cellcolor{top1}100.0   
&37.5   &  \cellcolor{top3}58.6 & \cellcolor{top1}100.0 
& \cellcolor{top3}49.1 &\cellcolor{top3}47.2 & \cellcolor{top1}100.0        \\

$\mathrm{BEF_{PGD}}$ (ours)     
&10.0  &12.2  &90.3  
&7.7  &11.0  &96.5  
&8.6  &15.2  &85.7\\

$\mathrm{CBEF_{PGD}}$ (ours)    
&\cellcolor{top3}48.8  & \cellcolor{top1}52.3  & \cellcolor{top1}100.0  
&\cellcolor{top1}54.4  &\cellcolor{top2}71.9  &\cellcolor{top1}100.0  
&\cellcolor{top1}61.5  &\cellcolor{top1}66.7  &\cellcolor{top1}100.0\\

\bottomrule

\end{tabular}
}
}
\label{tab:transferabilty_3000}
\end{table*}

\subsubsection{Quantitative Analysis on Transferability}
\label{subsubsec:transferability}



Transferability refers to the capability of an adversary in attacking one target model with the adversarial examples crafted from another model.
It is important to evaluate the transferability as it indicates the potential ability of a model in achieving effective blackbox attacks, which are more consistent with real-world attack problems.
For a fair comparison, we conduct the transferability experiment with adversarial examples of similar image quality.
After evaluating the performance of all baselines and our proposed methods in terms of SSIM and BRISQUE in Fig.~\ref{fig:exp_whitebox}, we have decided to use the following hyperparameters: a value of $32$ for all additive-perturbation-based baselines, a step size of $0.01$ for our proposed methods, an adversarial loss weight $c$ of $1.0$ for the C\&W attack, a maximum number of perturbation pixels $k$ of \revised{$1 \times 10^4$} for the SA attack.
%

We first craft adversarial examples from three models, \revised{for example}, ResNet50, MobileNet, and EfficientNet, respectively, and feed them to each model for evaluation. Finally, we obtain nine attack success rates for each attack method, including three whitebox attacks and six transfer attacks. Here, whitebox attack means attacking the target model with the adversarial examples crafted from the model itself. In contrast, the transfer attack fools the target model with the adversarial examples crafted from another model. 
\revised{
The experimental process is detailed in Algorithm~\ref{alg_quanti_transfer}.
}

Table~\ref{tab:transferabilty_3000} shows the results. 
As can be seen, most baselines as well as our methods (\revised{}BEF, CBEF, \BEFPGD{}, and \CBEFPGD{}), achieve almost $100\%$ success rate when they implement a whitebox attack. In addition, we have the following observations on transferability: 
\ding{182} BEF shows a significant advantage in image quality, but sacrifices its transferability. 
%
\revisedd{While BEF excels in image quality, it compromises on transferability. In white-box attacks from MobileNet, BEF and CBEF achieve high success rates of $95.4\%$ and $100\%$. However, when transferred to ResNet50 and EfficientNet, BEF's success rates drop to $8.8\%$ and $14.8\%$, which are even lower than MIFGSM. Similar results are observed with $\mathrm{BEF_{PGD}}$.
}
\ding{183} 
%
%
%
%
\revisedd{
CBEF improves transferability at the cost of image quality compared to BEF. For attacks on EfficientNet using adversarial examples from MobileNet, CBEF achieves a success rate of $58.6\%$, the third-best among all methods. Its PGD-enhanced version, \CBEFPGD{}, reaches $71.9\%$, while VNIFGSM leads with $81.5\%$. On ResNet50, \CBEFPGD{} ranks first with a $54.4\%$ success rate.
}
\ding{184} 
The transferability of CBEF is significantly increased by PGD. Concretely, take the adversarial outputs crafted from EfficientNet for example, its transfer attack success rate against ResNet50 is $61.5\%$, beating the $49.1\%$ of CBEF and $54.7\%$ of IFGSM.
We obtain similar results for the other models.

\subsubsection{Qualitative Analysis}
\label{DR_exp_QualiA}
%
To illustrate the advantage of our method in detail, we provide five visualization examples in Fig.~\ref{fig:via_baselines}. 
Specifically, after conducting the attack on the ResNet50 model as described in Sec.~\ref{subsubsec:QuantitativeAna}, we present the adversarial examples generated by our methods alongside those from the baselines. The adversarial patterns and perturbation levels of attack methods are visualized and compared in this section. We have the following observations:

%
\revisedd{
\ding{182} Overall, the C\&W attack, being optimization-based and focused on minimizing distortion, usually produces the best image quality. However, BEF and \BEFPGD{} can outperform it in some cases. Both BEF and \BEFPGD{} create adversarial examples with significantly better image quality than other baseline methods because they do not show visible patterns. Additionally, \CBEFPGD{} and CBEF also offer better imperceptibility than most baselines.
}
%
\revisedd{
For the first case, TIIFGSM and TIMIFGSM both achieve an SSIM value of $0.807$, followed by SA with $0.803$. In contrast, our BEF and \BEFPGD{} attain higher values of $0.966$ and $0.994$, respectively. CBEF, however, only achieves $0.743$. For the fourth case, CBEF performs better, with an SSIM value of $0.912$, surpassing the best additive-perturbation-based result of $0.736$ by TIIFGSM.
}
%

%
\revisedd{
\ding{183} In addition, the Laplacian pyramid fusion process in our framework clearly minimizes the noise-like patterns that are typically present in additive-perturbation-based baselines. As a result, the smoothness of the outputs allows us to achieve the best (lowest) $L_2$ and BRISQUE values.
}
\revisedd{
Upon evaluating the $L_2$ norm metric, \BEFPGD{} achieves the highest image quality among our attacks, with a value of $2.1 \times 10^{-6}$ in case 2, outperforming even C\&W's $5.6 \times 10^{-6}$. \CBEFPGD{} also performs well, benefiting from the PGD algorithm. BEF and CBEF yield $L_2$ norm values ranging from $10^{-3}$ to $10^{-4}$, comparable to other additive-perturbation-based attacks and SA.
}
%
\revisedd{
For the last case, TIIFGSM and TIMIFGSM both achieve BRISQUE scores of $37.39$, with JITTER at $49.75$, making them the top additive-perturbation-based baselines. However, our BEF method surpasses them with a significantly better score of $14.10$, even outperforming C\&W’s score of $15.35$.
}

%

\ding{184} Moreover, our method benefits from PGD in terms of image quality. Concretely, not only the \BEFPGD{} has a better SSIM value of $0.994$ than BEF ($0.966$) in the first case, the \CBEFPGD{} also generates adversarial examples with a higher SSIM value of $0.926$ than CBEF ($0.743$).
%

%

\revisedd{
\ding{185} At last, it is easy to see that our method's adversarial examples for case 2 have better image quality when compared to other cases. One possible reason is that the input image of case 2 is captured under unfavorable exposure conditions in comparison to other cases, making it easier to attack with minimal distortion. This phenomenon further demonstrates the threat of exposure changes to DNN-based DR grading system.
}

\begin{table*}[t]
\centering
\small
\caption{Adversarial comparison results on EyePACS~2015 dataset. We show the success rates ($\%$) of transfer \& whitebox adversarial attacks on three fine-tuned models: ResNet50, MobileNet, and EfficientNet, using our methods, \revised{which are} BEF and CBEF, with different numbers of Laplacian-pyramid levels ($L=1, 3, 5$). The last column in each three columns shows the whitebox attack results. The first and second columns show the transfer attack results.
}

{
\resizebox{.9\linewidth}{!}{ 
\begin{tabular}{l|ccc|ccc|ccc}

\toprule

\rowcolor{tabgray}\multicolumn{1}{c|}{Crafted from} & \multicolumn{3}{c|}{ResNet50} & \multicolumn{3}{c|}{MobileNet} & \multicolumn{3}{c}{EfficientNet} \\

\cmidrule(r){1-1} \cmidrule(r){2-4} \cmidrule(r){5-7} \cmidrule(r){8-10}
\rowcolor{tabgray} Attacked model
&  MobileNet   &   EfficientNet &  ResNet50

&  ResNet50   &   EfficientNet &  MobileNet

&  ResNet50      &  MobileNet &   EfficientNet\\

\midrule
BEF $(L=1)$      &10.1  &12.1 &99.7   &9.9   &10.5 &100.0&11.1 & 11.3 & 99.8   \\
BEF $(L=3)$      &11.3  &13.4 &89.2   &8.8   &14.8 &95.4 &9.5  & 17.5 & 84.2   \\
BEF $(L=5)$       &8.4  &10.3 &71.8   &6.3   &8.1  &79.7 &4.3  & 7.2  & 51.6   \\
\midrule
CBEF $(L=1)$      &12.1  &13.7 &100.0   &13    &21.3 &100.0  &10.5 & 10.9 & 100.0    \\
CBEF $(L=3)$      &27.6  &29.7 &100.0   &37.5  &58.6 &100.0  &49.1 & 47.2 & 100.0    \\
CBEF $(L=5)$      &34.4  &23.2 &100.0   &19.4  &39   &100.0  &43   & 34.3 & 100.0    \\

\bottomrule

\end{tabular}
}
}
\label{tab:pyramid}
\end{table*}

\subsection{Ablation study}
\label{subsec:ablation}
In this section, we further study the influence of the various hyperparameters introduced in Sec.~\ref{sec:method} on the attack success rate and transferability, including the level $L$ of the Laplacian-pyramid space in the adversarial bracketed exposure fusion as well as the kernel size $K$ in the convolutional bracketed exposure fusion.

\begin{table*}[ht]
\centering
\small

\caption{Adversarial comparison results on EyePACS~2015 dataset. We show the success rates ($\%$) of transfer \& whitebox adversarial attacks among three fine-tuned models: ResNet50, MobileNet, and EfficientNet, using our method with different numbers of pyramid levels ($L=1, 3$) and kernel sizes ($K=1, 3, 5$). The last column in each three columns shows the whitebox attack results. The first and second columns show the transfer attack results.}

{
\resizebox{.9\linewidth}{!}{ 
\begin{tabular}{l|ccc|ccc|ccc}

\toprule

\rowcolor{tabgray}\multicolumn{1}{c|}{Crafted from} & \multicolumn{3}{c|}{ResNet50} & \multicolumn{3}{c|}{MobileNet} & \multicolumn{3}{c}{EfficientNet} \\

\cmidrule(r){1-1} \cmidrule(r){2-4} \cmidrule(r){5-7} \cmidrule(r){8-10}
($L$, $K$)
&  MobileNet   &   EfficientNet &  ResNet50

&  ResNet50   &   EfficientNet &  MobileNet

&  ResNet50      &  MobileNet &   EfficientNet\\

\midrule

(1, 1)      &10.1 &12.1   &99.7 &9.9  &10.5 &100.0 &11.1 &11.3 &99.8        \\
(1, 3)      &12.1  &13.7 &100.0 &13   &21.3   &100.0  &10.5 &10.9 &100.0       \\
(1, 5)      &16.0  &17.6 &100.0   &19.4   &32.5 &100.0 &17.8 &23 &100.0        \\

(3, 1)      &11.3  &13.4 &89.2   &8.8   &14.8 &95.4 &9.5 &17.5 & 84.2        \\
(3, 3)      &27.6  &29.7 &100.0   &37.5   &58.6 &100.0 &49.1 &47.2 &100.0        \\
(3, 5)      &23.9  &24.0 &100.0   &37.8   &46.1 &100.0 &49.1 &49.7 &100.0        \\
\bottomrule

\end{tabular}
}
}

\label{tab:kernelsize}
\end{table*}

\subsubsection{Effects of the Number of Pyramid Levels \texorpdfstring{$L$}.}
\label{subsubsec:ablationL} 
As mentioned in Sec.~\ref{sec:method}, we introduce the  Laplacian-pyramid space to overcome the noise patterns generated by the simple fusion in Eq.~\eqref{eq:expmodel}. Here, we tune the pyramid level $L$ in Eq.~\eqref{eq:bracketed_reconstruction} and Eq.~\eqref{eq:conv_bracketed_reconstruction} to check its influence on the attack success rate of BEF and CBEF. Specifically, we set $L=1, 3, 5$ in this experiment and generate three different versions of BEF and CBEF, respectively. Note that, $L=1$ refers to no pyramid fusion process being applied. The results are shown in Table~\ref{tab:pyramid}.

%
%


\revisedd{
First, both whitebox and transfer attacks under BEF weaken as $L$ increases. For example, in MobileNet, the whitebox attack success rate is $100\%$ when there is no pyramid fusion ($L=1$). However, as $L$ increases, the success rate decreases to $95.4\%$ and then to$ 79.7\%$. Similarly, the transfer attack success rate to EfficientNet rises from $10.5\%$ to $14.8\%$, but then drops to $8.1\%$ with higher $L$ values.
Second, while the whitebox attack success rates of CBEF remain at $100\%$ across all $L$ settings, its transfer attacks are also influenced by $L$. Like BEF, CBEF achieves the highest transfer attack success rates when $L=3$ in most cases.
These observations suggest that pyramid fusion negatively affects attack success rates. Although it improves image quality, it reduces the attack capability.
}


\begin{figure}[h!]
    \centering
    \includegraphics[width=.75\linewidth]{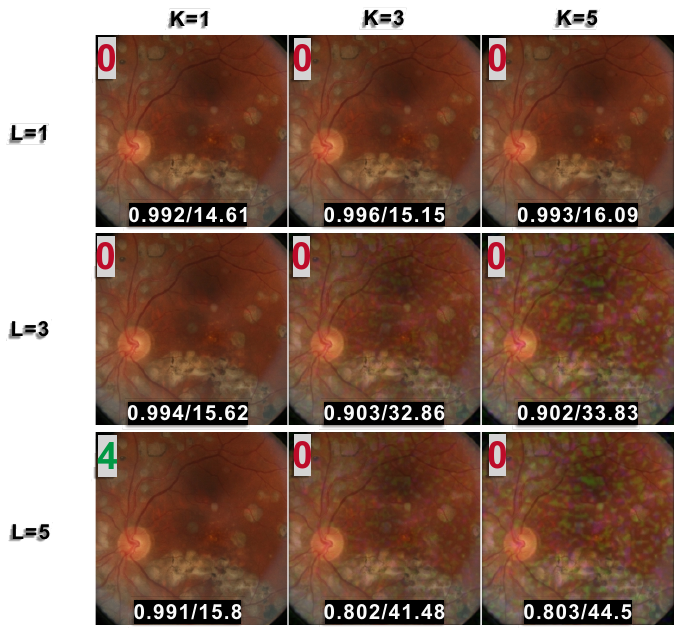}
    \caption{Visualization results of adversarial examples crafted from the ResNet50, using our methods, with different settings of L and K ($1$, $3$ and $5$). For each image, the DR grading result for ResNet50 is displayed on the top-left. The clean input is graded as $4$. The two numbers at the bottom are the SSIM and BRISQUE values.}
\label{fig:abl_via_baselines}
\end{figure}

\subsubsection{Effects of the Kernel Size \texorpdfstring{$K$}.}
\label{subsubsec:ablationK} 
We have shown the advantages of the convolutional bracketed exposure fusion in improving the transferability.
Here, we further conduct an experiment to study the influence of the kernel size $\mathbf{K}$ in Eq.~\eqref{eq:conv_bracketed_reconstruction} on the transferability of CBEF under different pyramid settings. 
More specifically, we pick two pyramid fusion levels, \revised{which are} $L=1, 3$, and tune the kernel size $K=1, 3, 5$.
The results are shown in Table.~\ref{tab:kernelsize}.

%
%
%
%

\revisedd{
When $L=1$, the transfer attack success rate increases as the kernel size $K$ grows. For adversarial examples generated from MobileNet, the success rate of transfer attacks on EfficientNet increases from $10.6\%$ to $32.5\%$.
When $L=3$, the transfer attack success rates increase with $K$ from 1 to 3 but then decrease as $K$ grows from 3 to 5. Again, for adversarial examples from MobileNet, the success rates on EfficientNet are $14.8\%$, $58.6\%$, and $46.1\%$ for $K=1$, $K=3$, and $K=5$, respectively.
For a fixed kernel size, such as $K=3$, the success rates of transfer attacks show a significant increase. Similar trends are observed for $K=5$. However, when $K=1$, which means CBEF simplifies to BEF, the attack success rate drops rapidly.
}

Overall, both the kernel size $K$ and the number of the pyramid level $L$ affect the success rate of transfer attack significantly.



\subsubsection{Visualization Results for Different Settings of \texorpdfstring{$K$,}~ \texorpdfstring{$L$}.}
\label{subsubsec:ablationvisual} 

To visually illustrate the influence of the hyperparameters, we show some examples in Fig.~\ref{fig:abl_via_baselines}, which are produced by different settings of $\mathbf{L}$ and $\mathbf{K}$.
%

%
%

\revisedd{
First, the LP fusion process counteracts noise in the additive attack (when $L=1$ and $K=1$). As $L$ increases to 3, the SSIM value improves from $0.992$ to $0.994$. However, when $K=3$ or $5$, where noise is minimal, LP fusion negatively affects image quality.
Additionally, a higher $L$ reduces the effectiveness of whitebox attacks. In the first column, the model is successfully fooled with $L=1$ and $L=3$ (graded as $0$) but fails with $L=5$ (graded as $4$).
}
\revisedd{
Second, a higher $K$ increases the whitebox attack capability but reduces image quality. When $L=5$, the method fails to attack with $K=1$ (graded as $4$) but successfully misleads the prediction to $0$ when $K=3$ and $K=5$. However, this also decreases the SSIM value from $0.991$ to $0.803$.
Moreover, a higher $K$ can weaken the noise pattern. In the first row, increasing $K$ from $1$ to $3$ improves the SSIM value from $0.992$ to $0.996$.
}

\subsection{Robustness Enhancement via Adversarial Training}
\label{DR_subsec:Advtrain}
As an attack, we can embed our methods to well-designed robustness enhancement algorithms to make DNNs robust to the exposure variation. 
Adversarial training \cite{athalye2018obfuscated} is one of the most effective defending methods, which generates adversarial examples at each minibatch during the training process and uses these examples as augmentations to update parameters of the DNN.
To validate the benefits of our attacks to the exposure robustness of DNNs, we implement the adversarial training technique with BEF to retrain the ResNet50 and obtain a robustness model denoted as ResNet50$_\text{at}$ that can resist the exposure variations. 
Note that, the full training dataset (35126 images) of EyePACS is involved. We can use adversarial exposure examples crafted from ResNet50$_\text{at}$, MobileNet, and EfficientNet to attack ResNet50$_\text{at}$ and validate the robustness enhancement.

%
%
%
%
%
\revisedd{
\paragraph{\textbf{Results}} The results in Table.~\ref{DR_Tab-advtraining} lead to the following observations:
\ding{182} Adversarial training with BEF improves the robustness of ResNet50 against exposure variations created by different DNNs (ResNet50, MobileNet, EfficientNet). This is indicated by the lower success rate of attacks on ResNet50$\text{at}$ compared to the original ResNet50. Specifically, under a whitebox attack, BEF has an $89.2\%$ success rate against ResNet50, but this drops to $6.6\%$ when attacking the enhanced model, ResNet50$\text{at}$.
Similar findings are observed in transfer attacks. For example, adversarial examples generated from EfficientNet have a $9.5\%$ success rate against ResNet50, which decreases to $2.3\%$ when targeting ResNet50$_\text{at}$.
\ding{183} The BEF-enhanced model (ResNet50$\text{at}$) remains effective under the CBEF attack. When using adversarial examples from MobileNet to attack both ResNet50 and ResNet50$\text{at}$, the success rate drops from $37.5\%$ to $21.6\%$, showing that the adversarial training-enhanced model can generalize to other types of attacks.
}

\begin{figure}
    \centering
    \includegraphics[width=.7\linewidth]{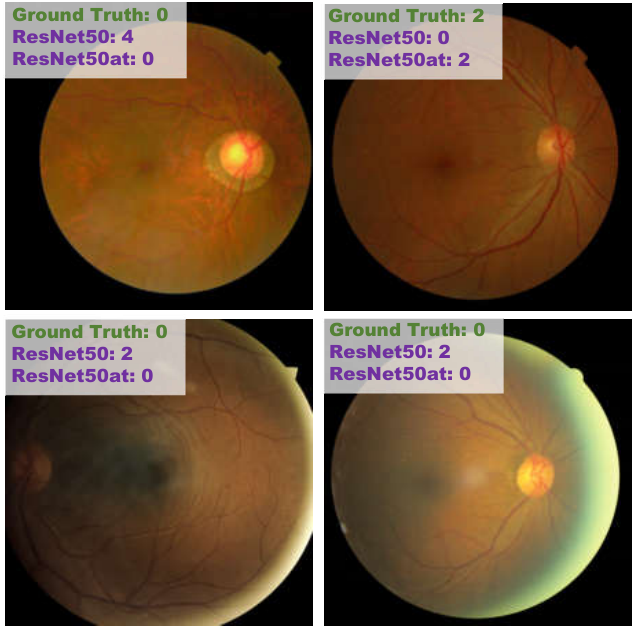}
    \caption{Recognition visualization results of four real RFIs from EyePACS. The ground truth labels are listed in the top-left of the input images. For each images, the predictions of original ResNet50 and BEF-enhanced model (ResNet50$_\text{at}$) are also listed in the top-left.}
\label{fig:realcase}
\end{figure}

\paragraph{\textbf{Visualizations}} To assess the robustness of the adversarially trained model against exposure in real-world scenarios, we present four real cases from the EyePACS dataset in Fig.~\ref{fig:realcase} that deceive the original ResNet50 model.
%
%
\revisedd{
All input images are captured under poor exposure conditions. The first two images in the top row are too dark for the DR grading model, while the remaining two suffer from light leakage on the right, causing uneven exposure. Despite these issues, doctors include these images in the dataset and accurately label them because the exposure problems do not affect their diagnosis. However, our experiments show that such conditions do impact the model's predictions. Our BEF-enhanced model, ResNet50$_\text{at}$, successfully predicts the DR severity for these poorly exposed images, indicating that our proposed method improves the model's robustness against exposure-related issues.
}

\begin{table}[t]
\centering
\small
\caption{Adversarial comparison results on EyePACS. It contains the success rates (\%) of our BEF and CBEF against the adversarially trained ResNet50 (ResNet50$_\text{at}$).
}
\label{DR_Tab-advtraining}
\begin{tabular}{l|c|c|c}
\toprule
\rowcolor{tabgray}{Crafted from} & Attacked model & \multicolumn{2}{c}{Attack methods} \\
\rowcolor{tabgray}{} & {} & BEF & CBEF\\
\midrule

ResNet50$_\text{at}$  &                      &6.6    &100.0   \\
MobileNet             &ResNet50$_\text{at}$  &2.4    &21.6    \\                 
EfficientNet          &                      &2.3    &29.6     \\
\midrule
ResNet50              &                      &89.2   &100.0    \\
MobileNet             &ResNet50              &8.8    &37.5    \\                 
EfficientNet          &                      &9.5    &49.1     \\

\bottomrule
\end{tabular}
\end{table}

\section{Conclusion}\label{sec:concl}
\revisedd{The motivation of this paper is to explore how camera exposure issues during the capture of RFIs affect DR diagnostic models. 
To address this, we propose a novel adversarial attack method, the \textit{adversarial exposure attack}, to highlight the potential threat that camera exposure poses to automated DNN-based DR diagnostic systems.
}
%
Our main contributions are summarized as follow: 
we first demonstrated challenges of this new task through a straightforward method, \textit{multiplicative-perturbation-based exposure attack}, where the naturalness of the exposure \revised{(inherent smoothness)} cannot be maintained. 
Then, we proposed the \textit{adversarial bracketed exposure fusion} based attack, in which the attack is formulated as a fusion of bracketed exposure sequences. 
Moreover, we further proposed an enhanced version by extending the multiplicative fusion to a convolution operation, which achieved significantly higher success rates for the transfer attacks. 
We have validated our method on a real and popular DR detection dataset, demonstrating that it can generate high-quality adversarial examples with high success rates for transfer attacks and help enhance the robustness of DNNs to exposure variations.

\revisedd{In this study, we explored the impact of exposure on DR models that accept RFIs as input. In the future, we plan to apply adversarial attack techniques to investigate model vulnerabilities across a broader range of input types, including multimodal problems. For example, COVID-19 diagnosis models use both X-ray images and CT scans as inputs \cite{9315478}. This will extend the framework presented in this paper to address more diverse challenges and optimize the corresponding models.}

\section*{References}
\bibliographystyle{IEEEtran}
\bibliography{ref}


\end{document}